\documentclass[
]{ceurart}

\usepackage{listings}

\usepackage{latexsym}
\usepackage{lmodern}
\usepackage{amssymb}
\usepackage{amsmath}
\usepackage{amsthm}
\usepackage{booktabs}
\usepackage{enumitem}
\usepackage{graphicx}
\usepackage{tabularx}
\usepackage{makecell}
\usepackage{svg}
\usepackage{xcolor}
\usepackage{float}
\usepackage{subfig}
\usepackage{soul}

\begin{document}

\copyrightyear{2024}
\copyrightclause{Copyright for this paper by its authors.
  Use permitted under Creative Commons License Attribution 4.0
  International (CC BY 4.0).}

\conference{CREAI 2024: Workshop on Artificial Intelligence and Creativity,
  Santiago de Compostela (Spain), 19-24 October, 2024}

\title{The creative psychometric item generator: a framework for item generation and validation using large language models}


\author[1]{Antonio Laverghetta Jr.}[
email=aml7990@psu.edu,
]
\address[1]{Department of Psychology, The Pennsylvania State University, 201 Old Main, University Park, Pennsylvania, USA}

\author[1]{Simone Luchini}[%
]

\author[2]{Averie Linell}[%
]
\address[2]{Department of Psychology, University of Nebraska at Omaha, 6001 Dodge Street, Omaha, Nebraska, USA}

\author[2]{Roni Reiter-Palmon}

\author[1]{Roger Beaty}

\begin{abstract}
  Increasingly, large language models (LLMs) are being used to automate workplace processes requiring a high degree of creativity. While much prior work has examined the creativity of LLMs, there has been little research on whether they can generate valid creativity assessments for humans despite the increasingly central role of creativity in modern economies. We develop a psychometrically inspired framework for creating test items (questions) for a classic free-response creativity test: the creative problem-solving (CPS) task. Our framework, the creative psychometric item generator (\texttt{CPIG}), uses a mixture of LLM-based item generators and evaluators to iteratively develop new prompts for writing CPS items, such that items from later iterations will elicit more creative responses from test takers. We find strong empirical evidence that \texttt{CPIG} generates valid and reliable items and that this effect is not attributable to known biases in the evaluation process. Our findings have implications for employing LLMs to automatically generate valid and reliable creativity tests for humans and AI.
\end{abstract}


\maketitle

\section{Introduction}

\begin{figure*}[h]
    \centering
    \includegraphics[width=0.8\linewidth]{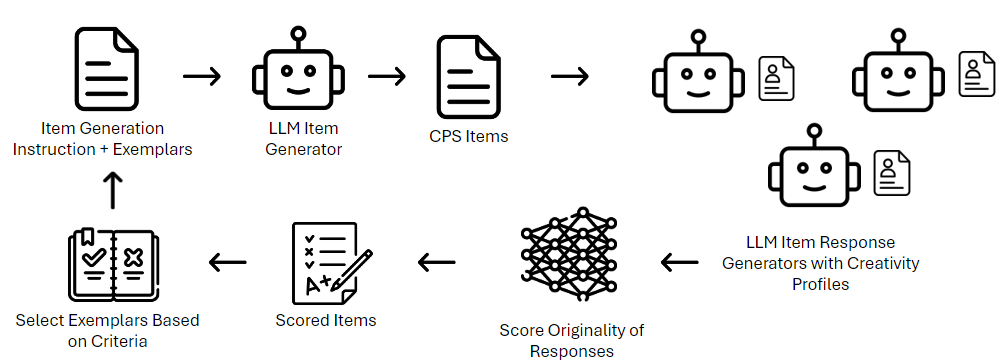}
    \caption{Overview of \texttt{CPIG}. From a base instruction, we prompt an LLM to generate CPS items, which are, in turn, completed by other LLMs. We give each LLM response generator a distinct profile to increase variability in the originality of their solutions. These responses are scored with an originality model developed by \cite{luchini2023automatic}, and a subset of the generated items with highly original responses are selected to include in the prompt for the next round of item generation. This figure was designed using images from Flaticon.com.}
    \label{fig:pipeline}
\end{figure*}

Creativity is considered one of the primary factors that determine individual \cite{mako2020automation} and organizational \cite{tsegaye2019antecedent} success in the modern economy. This is due to improved automation of routine tasks \cite{chui2015four}, the increasing complexity and ambiguity of problems organizations face, and projected growth of the creative sectors of the economy \cite{amabile2020creativity}. As such, the development of validated creativity tests has become increasingly important. Nevertheless, generating new creativity assessments remains a resource-intensive process requiring many hours of trial and error to develop suitable items (questions). Such items can be highly complex, requiring participants to reason about intricate scenarios or design solutions to ambiguous problems \cite{luchini2023automatic}, and therefore are difficult for even subject matter experts to develop.

With the introduction of modern large language models (LLMs) \cite{bommasani2021opportunities,brown2020language} the ability of AI to automatically develop novel creativity tests appears increasingly plausible \cite{rafner2023creativity}, and LLMs are already being used to automatically generate items measuring a variety of cognitive skills \cite{von2024item,lee2023paradigm,laverghetta-jr-licato-2023-generating}. Applying similar ideas in creativity assessment could provide a method to generate valid and reliable creativity tests at scale, which would be beneficial for assessing creativity in both humans and AI. However, doing so may also be contentious for some, given the broader debate on whether AI can be creative. Despite some evidence pointing towards AI creativity, whether AI-generated ideas are truly novel remains a hotly debated topic \cite{saebo2024stochastics,franceschelli2023creativity}. Some research suggests that using LLMs may lower the diversity of ideas produced over time, resulting in reduced collective novelty \cite{anderson2024homogenization,doshi2023generative}. Public perception of the creativity of AI also remains mixed; humans tend to view creative works produced by AI as less novel than those produced by other humans \cite{anderson2024homogenization}, and this could be problematic if humans become aware that they are being given AI-generated creativity tests. Broader research in social psychology has found that LLMs produce highly similar responses to questions regarding political orientation, moral philosophy, and other complex constructs that usually exhibit high variability in humans \cite{park2024diminished}. Collectively, these results point to a diminished diversity of thought in LLMs, which has important implications for whether and how LLMs should be used to automate creativity assessment.
 
How can we employ LLMs in designing items for measuring creativity without comprising the validity of any conclusions drawn from such items? We approach this from a \textit{psychometric} perspective, which is both a field dedicated to measuring psychological constructs in humans and the source of a rich body of work measuring similar constructs in AI \cite{attali2022interactive,vania2021comparing,laverghetta2021can}. When measuring a construct like creativity, psychometrics requires that any measurement be both valid and reliable --- it must accurately measure the intended construct and give consistent results over repeated measurements. Accomplishing this involves developing tests whose items accurately measure the construct, which historically was done by human experts. \ul{Can we use LLMs to generate high-quality items for measuring creativity?} If so, this would be invaluable not only for the study of human creativity but it might also allow us to measure creativity more accurately in LLMs, which would be a boon for assessing AI creativity. Nevertheless, no prior work has investigated whether LLMs can automatically generate creativity assessments.

In this paper, we develop a framework to extend item generation into the creativity domain: the \textit{creative psychometric item generator} (\texttt{CPIG}). \texttt{CPIG} relies on structured prompting and psychometrically based exemplar selection to generate items for a creative problem-solving task (CPS), an influential test of creativity \cite{reiter2009creativity}. Our framework is iterative and allows us to continuously refine the same item based on automated validity metrics until reaching a desired level of quality. While other works have explored how to use LLMs to solve \cite{rick2023supermind} and generate \cite{tian2023macgyver} CPS-like items, none to our knowledge has examined how to generate psychometrically rigorous assessments of creativity. We find that \texttt{CPIG} generated items are just as valid and reliable as those written by humans. Remarkably, LLM solutions to \texttt{CPIG} items also appear to become more original over successive rounds of generation, suggesting a possible method to boost the creativity of generative AI via carefully designed items.

We make the following contributions:
\begin{enumerate}
    \item We develop \texttt{CPIG}, a new framework for generating creativity items using LLMs.\footnote{Code and supplementary materials will be provided at: https://osf.io/umnk5/}
    \item Through a series of experiments, we confirm that \texttt{CPIG} generated items are just as valid as those written by humans, and that our metrics for validity are robust to known biases in the scoring process.
    
\end{enumerate}

\section{Background}
Creativity is thought to comprise multiple facets, including originality (the novelty of an idea) and effectiveness (how useful or relevant the idea is), among others \cite{diedrich2015creative}. Past work has demonstrated that human judgments of originality are an effective predictor of the creativity of ideas \cite{diedrich2015creative}. As such, the value of a creativity test rests on its capacity to elicit many original responses \cite{runco2012standard}. To measure originality, researchers historically relied on human judgments performed by trained raters --- a method called the Consensual Assessment Technique (CAT) \cite{silvia2008assessing}. In the CAT, human raters are instructed to read a series of ideas and assess their originality on a Likert scale. Although effective, human scoring is not efficient, as the recruitment and training of human raters is often costly and prone to errors. More recently, automated creativity assessment tools have been developed, including finetuning LLMs to predict human creativity ratings \cite{luchini2023automatic}. Highly accurate models have been reported, often matching or surpassing the agreement between human raters, which makes it practical to evaluate the quality of creative responses at scale.

From a psychometric perspective, measuring an individual's creativity requires developing structured tasks to evaluate how well they can produce ideas that are both original and high quality. We focus on a CPS as the basis for our experiments. In this task, a participant is given a scenario involving a dilemma to be solved (e.g., a coworker's roommate is causing problems at work, and it may put both of their jobs at risk), and they must produce a creative solution to this dilemma \cite{luchini2023automatic}. Scenarios are ambiguous by design, with many possible solutions, and reflect creative thinking in day-to-day settings. We focus on this CPS task due to its popularity as a creativity test and the availability of automated and psychometrically validated models for assessing the originality of CPS responses \cite{luchini2023automatic}. However, because many creative tasks can be evaluated in terms of originality, our methods are extensible to other tasks that can be automatically scored.  

\section{The architecture of \texttt{CPIG}}

We take a psychometric approach to generating CPS items, inspired by recent work on automatically generating psychometrically valid test items \cite{laverghetta-jr-licato-2023-generating, von2024item, attali2022interactive}. We use LLMs to act as \textit{item generators} to write the items, \textit{item response generators} to create human-like solutions to the items, and \textit{item scorers} to score the originality of LLM responses using psychometrically validated metrics. We hypothesize that originality in item responses provides a proxy for item quality: items with high quality should enable more creative responses and will tend to elicit better originality scores on average than those that are of lower quality. Optimizing for originality thus provides a way to generate higher quality items that can better tap the creative potential of subjects. Figure \ref{fig:pipeline} shows an overview of \texttt{CPIG}.

\subsection{Item generation}
Automatically generating valid CPS items is a non-trivial task, as the items must describe sufficently complex scenarios to allow a wide variety of responses while also being sufficiently ambiguous that no single solution is canonically more ``correct'' than the others. Furthermore, we also want scenarios to describe a wide range of situations to avoid generating an item pool revolving around a narrow range of topics. We thus develop a multi-stage prompting method.\footnote{All prompts used throughout \texttt{CPIG} are listed in the supplementary material.} 

First, before any runs of \texttt{CPIG}, we first prompt \textsc{gpt-3.5-turbo} to generate lists of words, where each list contains three names, a place, and an action (e.g., ``Mark'', ``beach'', ``Amy'', ``Lucas'', ``swimming''). The goal behind this step is to make the item generation task more concrete; rather than prompting the item generator LLMs to design scenarios without any additional context, we instead use the word lists as criteria that must be satisfied (e.g., the final scenario must contain all the names from the word list). This is meant to both simplify generation by breaking it down into multiple steps and help maximize diversity in scenario content by using different word lists to ensure no two item generation prompts are the same. We have \textsc{gpt-3.5-turbo} generate ten word lists at once to help eliminate redundant lists and query the model five times to generate 50 lists in total. We set the max number of tokens to 2048 and the temperature to $1.0$, leaving other parameters at their defaults. We use this process to generate lists covering a wide variety of semantic content that we manually checked to confirm they obeyed the specified format. We use these word lists throughout all trials of \texttt{CPIG}.

We use these word lists in the item generation prompt, where we instruct item generator LLMs to design CPS items using the contents of the word list provided. We provide LLMs with generation guidelines and examples of CPS items written by experts. For each trial, we attempt to generate one scenario for each word list. However, the generated items may fail basic validity checks for a variety of reasons, so to mitigate this, we develop a list of rules to drop generations that are likely low quality: 

\begin{enumerate}
    \item We compute item readability using Flesch's reading ease \cite{kincaid1975derivation} and drop scenarios with scores lower than 45 (considered very difficult to read). We note that this metric requires a minimum string length to compute, so we also require that scenarios be at least 140 tokens long. We use the NLTK word tokenizer to ensure a conistent token count.\footnote{https://www.nltk.org/api/nltk.tokenize.word\_tokenize} 
    \item From preliminary trials, we find that LLMs sometimes generate scenarios with priming effects, steering participants toward specific solutions. Examples of this include generating a list of possible solutions or setting up the scenario as a dichotomy (``Should I do \textit{X} or \textit{Y}?''). Based on the content of such scenarios, we developed a list of strings that indicate possible priming and drop scenarios that contain any such string. Specifically, we drop scenarios containing ``on the one hand,'' ``on the other hand,'' ``dilemma,'' ``must navigate,'' ``must decide,'' ``has to decide,'' and ``is torn between.'' We do not claim that this list is comprehensive, but we found that it eliminated most priming in generated scenarios.   
    \item To prevent LLMs from generating irrelevant content after the scenario, we instruct them to always generate ``I am finished with this scenario.'' at the end. We drop scenarios that lack this string.  
\end{enumerate}

\noindent Importantly, our goal behind this quality control was not to identify every possible error that might occur in the items, as we expect human experts will make the final decision for which items to include in a creativity assessment \cite{von2024item}. Rather, we use it to reduce the number of items that need to be examined by eliminating those that are unlikely to be valid. We attempt to generate a scenario a maximum of 10 times for each word list and drop the list if the LLM fails to generate a valid scenario on all attempts. We strip extra newlines and whitespace surrounding the scenario and text after the termination string (including the string itself).

\subsection{Item response generation}
Once we have LLM-generated items, we must evaluate whether they elicit creative responses. LLMs have proven adept at modeling psychometric data \cite{laverghetta2021can} and are competent as human simulacra for sociological modeling \cite{sun2024random}, so we use LLMs to generate synthetic responses to each item. A potential challenge here is that the item response generator LLMs may suggest similar solutions to the same item \cite{anderson2024homogenization}. We account for this by adopting several prompting styles meant to increase the variation in the LLM responses: a \textit{baseline} prompt where the LLM is asked to provide a creative solution to the item (with no further context), a \textit{demographic} prompt where the LLM is provided demographic data about a hypothetical participant that it is meant to simulate while responding (e.g., ``You are a Hispanic woman who works in real estate''), and a \textit{psychometric} prompt where we replace the prior demographic data with statements sourced from psychometric inventories strongly correlated with creative performance. 

For demographic and psychometric prompts, we construct a pool of \textit{participant creativity profiles} to draw from based on responses to prior creativity studies \cite{luchini2023automatic}. These responses include differing occupations and responses to psychometric assessments, which we reason would increase the variability in the output of the item response generator LLMs. We provide demographic data in the prompt using either a variable format (e.g., "You are an Asian man") or as demographically relevant names. Demographic variables, including name, ethnicity, and gender, were taken from the New York City Health Department 2016 census of baby names,\footnote{https://www.nyc.gov/site/doh/index.page} and last names specifically were taken from the Decennial Census Survey\footnote{https://www.census.gov/programs-surveys/decennial-census.html} from the United States Census Bureau. We selected the three most common first and last names associated with each demographic variable for a total of 20 first names and 20 last names. We extract data for the psychometric prompts from a series of validated scales measuring constructs related to creativity. We employed scales tapping creative self-efficacy \cite{karwowski2012did}, creativity anxiety \cite{daker2020creativity}, creative mindset \cite{karwowski2014creative}, openness to experience \cite{deyoung2014openness}, tolerance for ambiguity \cite{furnham1995tolerance}, cynicism \cite{mitchell2023malevolent}, and the RIASEC interest types \cite{armstrong2008holland}.

In each prompting style, the model is provided a CPS item after the task instructions and demographic/psychometric profile (if applicable), and we process the generated response by removing extra newlines and white space. Because response generation is comparatively a much simpler task than item generation, we do not include additional content validity checks. We generate between 10 to 20 responses for each item. For the demographic and psychometric prompts, we sample a participant profile at random each time.

\subsection{Item scoring and selection}
Each LLM-generated item response is then scored using the methodology developed by \cite{luchini2023automatic}, which trained \textsc{roberta-base} \cite{liu2019roberta} to predict mean originality scores of responses to CPS items. Specifically, this model was trained on a dataset annotated by experts for originality, who scored each response using a five-point Likert scale. They used a test set comprising originality scores to CPS items not seen during training and obtained a $0.41$ Pearson correlation with human ratings. We use this model to score the originality of each \texttt{CPIG} item, which we use to select $k$ items to include as exemplars in the next round of item generation. We develop several shot selection strategies for choosing exemplars, which we discuss below. Additionally, we include a baseline that simply chooses $k$ items at random.

\subsubsection{Greedy}
This approach simply selects the $k$ items with the highest originality scores. Specifically, we take the mean of the originality scores of all the responses per item and sort the resulting scores to select the $k$ items with the highest scores.

\subsubsection{Constraint satisfaction}
A challenge with the greedy approach is that it may choose highly similar items if they all score high on originality. Indeed, we found in preliminary trials that cosine similarity scores between all pairs of the $k$ items tend to increase over iterations, sometimes drastically. To address this, we develop another shot selection method that instead finds a set of $k$ items that maximize originality and minimize similarity, which we treat as a constraint satisfaction problem. For each iteration of \texttt{CPIG}, we have a set of exemplars $I$ from the prior iteration\footnote{We still employ the greedy approach for the first iteration, as we don't yet have values to compare against.} with a mean originality score $I_o$ and a mean semantic similarity $I_v$ (the mean cosine similarity scores between all pairs of items in $I$). Additionally, we include thresholds $\delta_o$ and $\delta_v$ that define a tolerance above $I_v$ and below $I_o$ for the new set of exemplars. We then search for a set $\eta$ of size $k$ from the generated item pool at the current iteration that satisfies: 

\begin{equation}
    \eta_o > I_o \; \lor \; I_o - \eta_o \le \delta_o
\end{equation}
\begin{equation}
    \eta_v < I_v \; \lor \; \eta_v - I_v \le \delta_v
\end{equation}

We use Sentence Transformers \cite{reimers-2019-sentence-bert} and \textsc{all-MiniLM-L6-v2} to compute $I_v$ and $\eta_v$, and we search for all matching $\eta$ across all unique combinations of size $k$ from the item pool. We return the $\eta$ with the highest originality score; further details on this method and the chosen values for $\delta$ are provided in the supplementary material.

\begin{figure}[htb]
    \centering
    \footnotesize
    \includegraphics[width=0.5\linewidth]{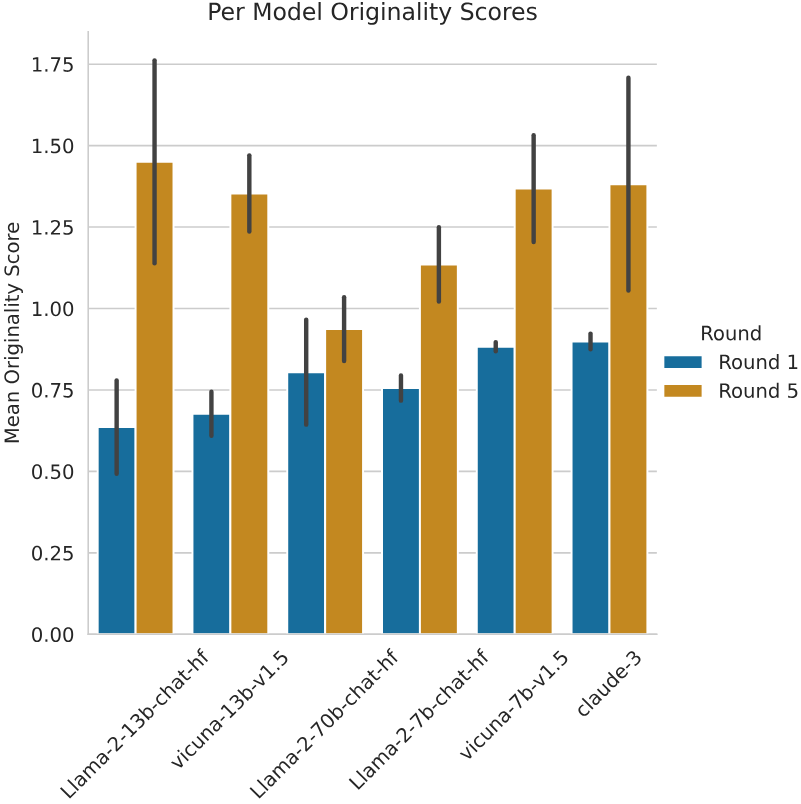}
    \caption{Mean originality scores from each item generator on the first and last rounds, for all trials that did not use random shot selection. Error bars are standard deviations in scores. Higher values indicate more original item responses, on average.}
    \label{fig:per-model-scores}
\end{figure}

\subsection{Implementation details}
We implement \texttt{CPIG} using LangChain\footnote{https://www.langchain.com/} and utilize a variety of chat-based open-source and commercial LLMs, including \textsc{LLama-2} (7b, 13b, and 70b) \cite{touvron2023llama}, \textsc{Vicuna-1.5} (7b and 13b) \cite{vicuna2023}, and \textsc{Claude-3-haiku}.\footnote{https://www.anthropic.com/news/claude-3-family} All open-source models are implemented using Transformers \cite{wolf-etal-2020-transformers}. We set the temperature to $1.0$ across all trials to increase variation in the generated items and responses while leaving other text generation parameters at their defaults. We select four items to use as exemplars for all shot selection methods to ensure item generation prompts do not become too long and because we find this is sufficient to ensure variation in item content. We cap item generation to a maximum of 768 tokens and item response generation to 350 tokens, as responses to CPS items tend to be much shorter than the items themselves. We run each \texttt{CPIG} trial for five iterations, using three random seeds for every hyperparameter combination. We use the same LLM for item generation and item response generation for each open-source model trial and use \textsc{LLama-7b} for response generation when using \textsc{Claude-3-haiku} for item generation. We provide a table listing all trials in the supplementary materials. We run experiments on three Nvidia RTX A6000 GPUs with 49GB of video memory each. We apply 4-bit quantization to all supported models.


\begin{figure}[h!]
    \centering
    \footnotesize
    \includegraphics[width=0.5\linewidth]{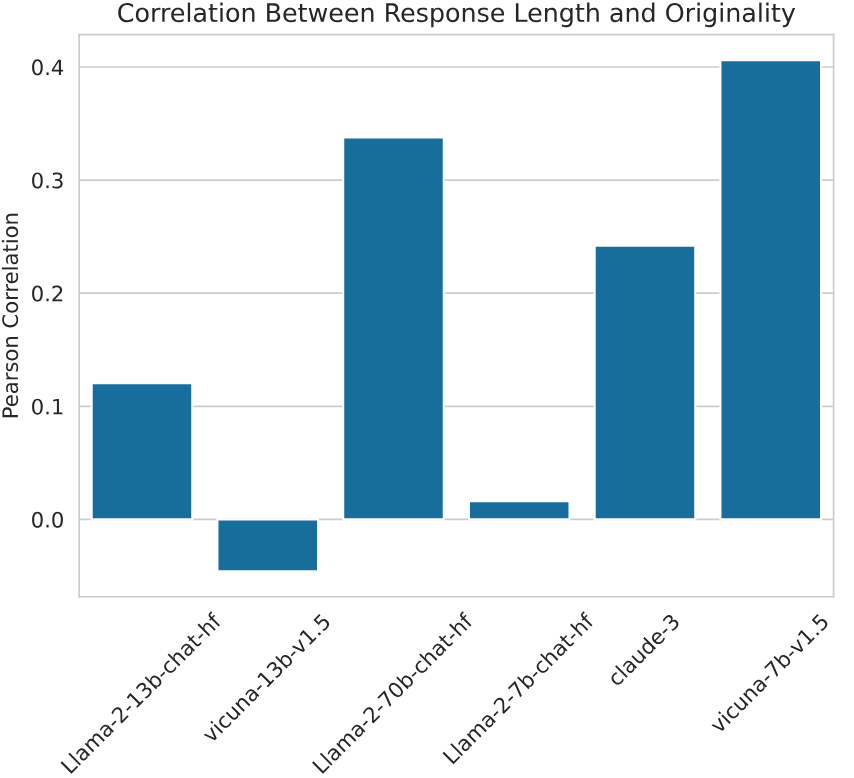}
    \caption{Pearson correlation between item response length and originality score. Length is calculated using the NLTK word tokenizer.}
    \label{fig:correlation}
\end{figure}

\section{Results}
We present a comprehensive picture of how effective the different components of \texttt{CPIG} are at generating items that maximize the originality of the output from item response generator LLMs. This includes both ablations on the effect of the different prompting strategies and shot selection methods, as well as human review on the quality of the generated items. For any ablation that requires computing semantic similarity, we use Sentence Transformers \cite{reimers-2019-sentence-bert} and \textsc{all-MiniLM-L6-v2} as the embedding model. All density plots employ kernel density estimation \cite{parzen1962estimation}.

\begin{figure}[h]
    \centering
    \includegraphics[width=0.5\linewidth]{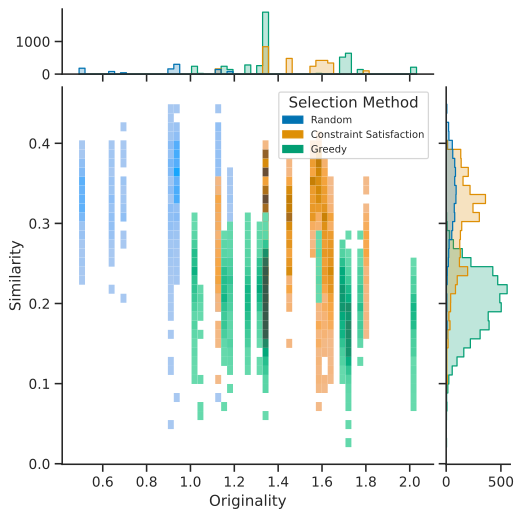}
    \caption{Joint histogram of originality and similarity scores for round five items. The highest quality items are those in the bottom right region. Note that we have dropped all items whose cosine similarity was greater than $0.95$ to any other item.}
    \label{fig:orig-scores}
\end{figure}

\subsection{Originality of LLM responses}
Figure \ref{fig:per-model-scores} shows originality scores for all runs that do not use random shot selection, broken down by model type. Critically, regardless of the item generator, \texttt{CPIG} consistently improves originality scores of responses by the last round of item generation, in some cases \textit{more than doubling} the score compared to the first round. The difference in mean scores was significant in $t$-tests for both demographic ($p << 0.001$) and psychometric ($p << 0.001$) prompting styles and hence remains regardless of the specific prompting strategy used for item response generation. This demonstrates that \texttt{CPIG}-generated items can elicit more creative responses from the item response generator LLMs. However, a potential confound when scoring originality is that the metric is influenced by the length of the response, with longer solutions typically being scored as more original \cite{luchini2023automatic}. We find that LLM responses are, on average, much longer than those of humans, leaving open the possibility that the increase in originality is driven purely by more elaboration in the response. We check for this by computing the Pearson correlation between response length and originality for every generation model and the items generated on the last round (not including random shot selection). Results are shown in Figure \ref{fig:correlation}. As expected, length is at least partially correlated with originality for all generation models, though there is significant variation in the strength of this correlation. Importantly, however, the correlations remain weak overall and do not rise above $0.3$ in either direction for most LLMs, suggesting that the increases in originality are not only due to increasing response length.

\begin{figure}[h!]
  \centering
  \subfloat[\centering Distributions of originality scores, broken down by item response prompting strategy. As a point of comparison, we also plot the originality scores of the human participants used to train the scoring model from \cite{luchini2023automatic}, but note that they are not given the same items generated by \texttt{CPIG}.]{\includegraphics[scale=0.38]{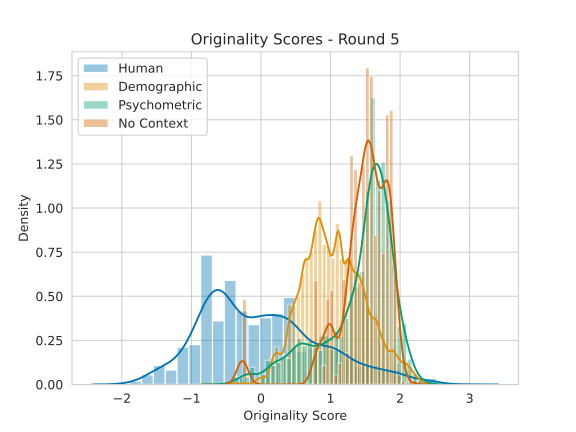}}
  \subfloat[\centering Cosine similarity scores between all pairs of items from the last round of generation, for both greedy shot selection and constraint satisfaction.]{\includegraphics[scale=0.38]{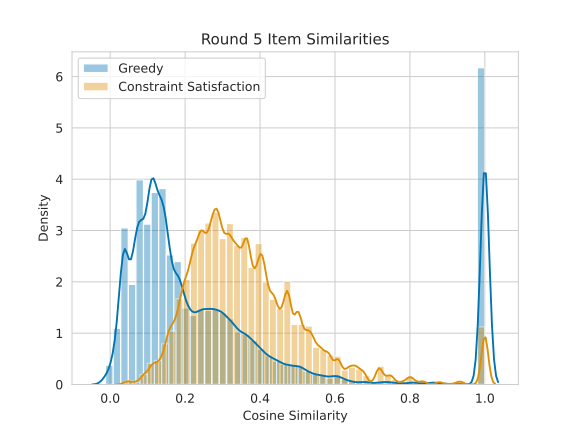}}
  \caption{Distributions of originality (a) and similarity (b) scores, broken down by prompt types and shot selection strategy.}
  \label{fig:item-sim-originality-scores}
\end{figure}

\subsection{Relationship between originality and similarity}
While improvements in response originality denote an increase in item quality, it remains unclear whether the item generator LLMs converge onto a few similar yet high-quality scenarios or how these variables relate to each other in the generated item pool. We explore this by plotting a joint histogram of originality and similarity scores\footnote{Measured as the mean cosine similarity between each item and every other item.} for all generated items, broken down by shot selection method, in Figure \ref{fig:orig-scores}. Darker cells in this figure indicate a higher frequency of a particular originality-similarity combination. We observe that random shot selection obtains the worst combination of results: not only are most items low on originality, but the distribution also peaks the highest on similarity. Both greedy shot selection and constraint satisfaction achieve lower similarity and higher originality and do so consistently. As the originality of items produced using these strategies increases, their similarity scores remain generally static, indicating that improvements in originality do not come at the expense of more redundant items.

One notable trend is that greedy shot selection seems to have lower similarity scores on average despite constraint satisfaction being designed to minimize similarity. However, for this figure, we dropped all items whose similarity is above $0.95$ to any other item to make computing the joint histogram more manageable. In Figure \ref{fig:item-sim-originality-scores}, we graph the univariate histogram of cosine similarity scores for both greedy and constraint satisfaction, and this time, include all the items that are generated in the last round. Although both methods generate some item pairs with cosine similarities of $1.0$, there are many more such items for greedy shot selection, indicating a much larger fraction of extremely similar item content. Interestingly, greedy also peaks at a higher density than constraint satisfaction towards the lower end of the distribution. This likely reflects the balancing act required for constraint satisfaction; selecting items to maximize originality may sometimes require increases in similarity, though the method still succeeds in eliminating most duplicate content.

\subsection{Effect of item response prompting style}
Humans typically exhibit high variability in the originality of their responses to CPS items \cite{luchini2023automatic}. The different item response prompting strategies we develop are meant to induce a similar degree of variation, and we examine how effective they are in Figure \ref{fig:item-sim-originality-scores}. Compared to the no-context baseline --- where the item response generator LLMs are simply instructed to answer the item --- both demographic and psychometric prompting strategies exhibit higher variance and heavier tails in the originality distribution, better reflecting the trends from human participants. Both curves still have lower variance than humans and much higher peaks in originality scores, so it appears there remains headroom for alignment between LLM and human psychometric properties. The main challenge here again relates to elaboration in the response; while human participants often give short solutions, LLMs tend to provide very elaborate responses that embed multiple solutions simultaneously. Fully overcoming this challenge requires more sophisticated prompting and perhaps additional finetuning on human responses to align with our preferences for this task, but we leave this to future work.

\subsection{Human content review}

\begin{figure*}
  \centering
  \subfloat[\centering Complexity]{\includegraphics[scale=0.39]{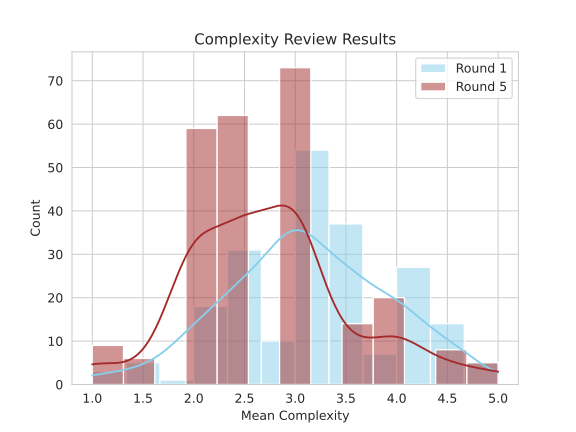}}
  \subfloat[\centering Difficulty]{\includegraphics[scale=0.39]{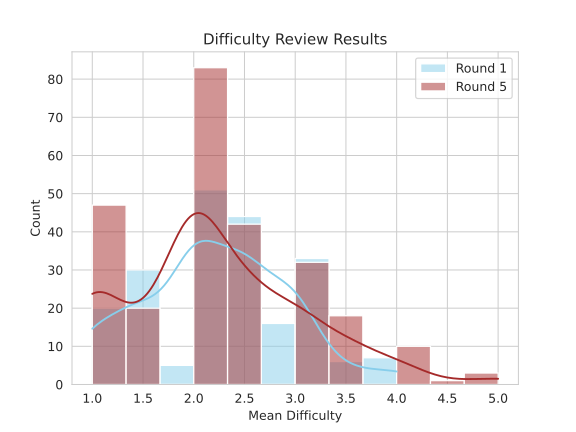}}
  \caption{Mean complexity and difficulty scores from round one compared against round five. A rating of three indicates ideal complexity/difficulty.}
  \label{fig:content_review}
\end{figure*}

The prior results demonstrate that, with carefully chosen prompts and few-shot exemplars, \texttt{CPIG} can generate items that elicit more original responses from LLM test takers. But is this trend due to improvements in item quality or some other artifact of the generation process? We explore this by recruiting human annotators to rate the quality of the \texttt{CPIG} items.   

We recruited five annotators with prior experience in rating for creativity studies. Annotators rated each item in terms of its \textit{complexity} and \textit{difficulty}, where we define complexity as how many \textit{demands} were present in the item and difficulty as how many of those demands directly compete with each other, such that a solution that attempts to solve one might come at the expense of another. We define demands as any relevant information in the scenario that could be used to construct a creative solution. Demands could include challenges to overcome in the scenario or resource constraints, among many others. We selected these facets to cover the most important factors to rate to ensure content validity in the items based on our expertise in creativity assessment and preliminary examinations of the items generated by \texttt{CPIG}. Both facets were rated on a five-point Likert scale, with one being too simple/easy, five being too complex/difficult, and three having the right amount of complexity/difficulty. This scale allowed us to account for both extremes of item content; items that are too complex or difficult might cause human participants to give up prematurely, while items that are too simplistic or easy are unlikely to require much creativity to solve. We designed a rubric that annotators used to rate each item, including definitions for complexity and difficulty. The annotators were first shown the rubric and allowed to ask any questions they had about the task. Then, together with one of the authors, the annotators rated ten practice items. Finally, the annotators, in combination with two of the authors, rated the remaining items via a missing data approach, where annotators only rated a subset of the \texttt{CPIG} items. This approach allowed us to achieve maximum coverage of all items while limiting rating time and making the annotation workload manageable. Each annotator rated between 200 and 245 LLM-written items, including items from the first and last round of \texttt{CPIG}. Annotators were only provided the text of each item, and were blinded to all other related details. For instance, annotators were not informed of which items belonged to which round of \texttt{CPIG}.

We obtained intraclass correlations of $0.52$ for complexity and $0.49$ for difficulty, for absolute agreement on the average ratings, indicating a modest rater agreement.\footnote{This was expected as rating creativity can be highly subjective, so it is challenging to achieve stronger rater agreement.} We plot in Figure \ref{fig:content_review} the distributions of complexity and difficulty scores from the items from the first and last rounds. For complexity, we see a definite improvement from round five, with a much larger fraction of items achieving the ideal complexity level than was present in round one. Trends are more static for difficulty as the distributions are quite similar to each other, especially at the ideal difficulty level. Collectively, the content review indicated that \texttt{CPIG} items are generally of high quality and that later iterations result in definite improvements for at least some facets of item quality.

We include two items generated by \textsc{LLama-13b} in Table \ref{tab:sample_items}, both using the same word list. While even items generated in the first round exhibit many desirable qualities, we see key improvements over iterations. Although the round one item (top row in the table) sets up what could be a complex scenario, it remains unclear what the exact problem is other than that Noah is being asked to do ``extra work'' for a customer. The round five scenario (bottom row) makes this clear: a new family is causing problems by stealing plants. This scenario also introduces added complexity by including new characters with interwoven relationships, hence adding more competing demands that need to be considered. The scenario is still not perfect as not all the information appears especially relevant, but overall, it does appear to be both more original and of higher quality.

\begin{table*}[h!]
\centering
\footnotesize
\caption{Example items generated by \textsc{LLama-13b} from the first (top row) and last (bottom row) iterations of item generation, using the same word list.}
\begin{tabular}{c}
CPS Item      \\ \hline                                                                                                                                                                                                                                                                                                                                                                                                                                                                                                                                                                                                                                                                                                                                                                                                                                                                                                                                                                                                                                                                                                                                                                                                                                                                                                \\
\begin{tabular}[c]{@{}c@{}}Noah is a 25-year-old man who works as a gardener at a local gardening company. \\ He is known for his excellent gardening skills  and is always willing to help his customers find what they need. \\ Recently, a new customer named Lily has started coming to the gardening company, \\ and she has been causing problems for Noah. \\ Lily is always asking Noah to do extra work for her, and she has been making him feel overwhelmed. \\ Noah is starting to feel frustrated and is unsure of what to do. \\ He does not want to lose his job, but he also does not want to work in an environment that is so tense and \\ hostile. He is considering talking to his manager about the situation, \\ but he is worried that it could backfire on him. He is unsure of what to do.\end{tabular} \\ \hline                                                                                                                                                                                                                                                                                                                                                                                                                                                                                        \\
\begin{tabular}[c]{@{}c@{}}Noah is a 35-year-old man who lives in a small town with his wife Lily and their 5-year-old son James. \\ Noah is a skilled gardener and spends most of his time tending to the plants in the community garden. \\ Lily is a painter and spends most of her time in her studio, but she also helps out in the garden \\ when she can. James loves spending time in the garden with his parents and is always eager to help out. \\ Recently, a new family moved into town and they have been causing problems for Noah and Lily. \\ The new family, the Smiths, have been stealing plants from the community garden and selling \\ them at the local farmer's market. Noah and Lily are not sure what to do about the situation. \\ They do not want to confront the Smiths directly, but  they also do not want to lose their plants. \\ They are considering asking James's teacher, Ms. Johnson, for help. Ms. Johnson is a kind and fair person, \\ but she is also a close friend of the Smiths. Noah and Lily are not sure if Ms. Johnson will be willing to help \\ them  or if she will be biased towards the Smiths. \\ They are also worried that if they do ask Ms. Johnson for help, it could cause problems for James in school. \\ They are at a loss for what to do.\end{tabular} \\ \hline
\end{tabular}
\label{tab:sample_items}
\end{table*}

\section{Related work}



\subsection{Psychometric AI}
Psychometric analysis of language models has seen growing interest in NLP research \cite{laverghetta-jr-licato-2023-generating,laverghetta2021can,laverghetta2021predicting,vania2021comparing,li2024quantifying,he2024psychometric}. Measurement models from psychometrics provide a strong test bed for evaluating language understanding in LLMs \cite{vania2021comparing}, making psychometrics a valuable tool for building better NLP test sets. However, LLMs are also valuable for modeling psychometric properties exhibited by humans on both cognitive \cite{laverghetta2021can} and non-cognitive \cite{lee2023paradigm} assessments, spurring interest in how LLMs might model human response data more broadly \cite{tavast2022language}. One rapidly growing research area is automated item generation, where LLMs are used to create new test items for standardized assessments with little or no human intervention \cite{von2024item,laverghetta-jr-licato-2023-generating}. Several works have proposed frameworks similar to ours, where multiple LLMs are used to iteratively generate and evaluate new test items \cite{hernandez2023ai,attali2022interactive}. However, this research has focused almost entirely on generating multiple-choice items, where the range of possible responses is inherently restricted. Additionally, the constructs targeted by such frameworks are either purely cognitive (with an objectively correct answer) or non-cognitive (open to interpretation based on individual differences). Creativity does not neatly fit into either mold: there is an aspect of ``correctness'' when judging CPS responses as the goal is to present a viable solution, yet how solutions are compared against each other in terms of originality is often open to rater interpretation \cite{benedek2013assessment}. Our work thus moves psychometric AI in a new direction to examine constructs outside the narrow scope explored in prior work.


\subsection{Prompt engineering for psychometric assessment}
An often-overlooked aspect of AI-based test development is prompt engineering: the process of developing prompts for LLMs that yield strong performance on the task of interest. Many studies rely on manual prompt tuning to adapt LLMs to a specific cognitive or psychometric task, which has allowed for the successful replication of many classic results from cognitive psychology \cite{ushio-etal-2021-bert} and has yielded high-quality items for various assessments \cite{lee2023paradigm}. A typical design pattern for such prompts is to use a format that aligns closely with how the actual task is presented to humans as if to simulate an experimental session \cite{tavast2022language}. However, greater care must be taken in the prompt design than might be necessary for other applications, as LLMs appear susceptible to more biases in task instructions than humans \cite{gupta2023investigating}. A starting point for addressing this could be to employ methods for prompt optimization, which have been widely successful in improving the performance of LLMs for NLP tasks \cite{zhou2022large}. These techniques, while powerful, typically rely on information-theoretic metrics for assessing prompt quality, often resulting in uninterruptible prompts \cite{liu2023pre}. A few works have explored how to create prompt optimization methods employing psychometrics as optimization targets by combining LLM item generators with discriminative models trained to predict item alignment with a target construct \cite{hernandez2023ai} or by incorporating standard metrics for reliability and validity to assess the quality of an LLM's generations \cite{laverghetta-jr-licato-2023-generating,attali2022interactive}. Even in these cases, the prompt itself usually remains static. \texttt{CPIG} provides a structured method for prompt mutation via the selection of exemplars that demonstrate evidence of validity on the task of interest.

\section{Conclusion}
We propose \texttt{CPIG}, a framework for generating creativity items using LLMs. By combining state-of-the-art models for response scoring with methods for item generation, we find that \texttt{CPIG} can generate items that improve the originality of LLM responses over time, which in turn points to increased creativity in their solutions. This trend is not attributable to known biases in the scoring model, and human raters find \texttt{CPIG} items to be high quality. 

While our results are promising, our analysis also has limitations. In developing \texttt{CPIG}, we focused primarily on originality as the metric to optimize. While originality is a crucial facet of creativity, it is just one metric for judging creative outputs. Depending on the context, other metrics, such as an output's quality or relevance, may be more important to evaluate, and future work should extend our framework to optimize multiple criteria simultaneously. The quality of the generated items depends directly on the item evaluation, which was accomplished through automated scoring that, while effective, is not without limitations \cite{luchini2023automatic}. Developing more robust evaluations requires layering multiple quality control checks on top of each other, perhaps by employing separate LLM judges to rate the quality of the items directly and provide structured feedback on how to improve the items. Though we performed a content review on the \texttt{CPIG} items, it remains unclear how effective they would be when administered to human participants to solve without conducting more studies. As such, we caution against using the items from \texttt{CPIG} until they have undergone more extensive review. Finally, we must acknowledge biases in the LLMs, which may have influenced item generation. The data for our scoring model was curated using raters from a Western background \cite{luchini2023automatic}, making the possibility of bias even more likely. Addressing this requires curating originality scores representing a more diverse slate of cultural views and developing bias mitigation strategies during item generation to ensure the evaluation remains fair.

\begin{acknowledgments}
  The research described herein was sponsored by the U.S. Army Research Institute for the Behavioral and Social Sciences, Department of the Army (Contract No. W911NF-23-C-0040 P00001). The views expressed in this article are those of the authors and do not reflect the official policy or position of the Department of the Army, DoD, or the U.S. Government. 
\end{acknowledgments}


\end{document}